\documentclass{article}

\usepackage[final]{neurips_2025}

\usepackage[utf8]{inputenc} % allow utf-8 input
\usepackage[T1]{fontenc}    % use 8-bit T1 fonts
\usepackage{hyperref}       % hyperlinks
\usepackage{url}            % simple URL typesetting
\usepackage{booktabs}       % professional-quality tables
\usepackage{amsfonts}       % blackboard math symbols
\usepackage{nicefrac}       % compact symbols for 1/2, etc.
\usepackage{microtype}      % microtypography
\usepackage{xcolor}         % colors

\usepackage{wrapfig}
\usepackage{graphicx}
\usepackage{subfigure}
\usepackage{multirow}
\usepackage{float}
\usepackage{amsmath,amsfonts,bm}

\def\eqref#1{equation~\ref{#1}}
\def\1{\bm{1}}

\def\vb{{\bm{b}}}

\def\vp{{\bm{p}}}
\def\vq{{\bm{q}}}

\def\vx{{\bm{x}}}
\def\vy{{\bm{y}}}
\def\vz{{\bm{z}}}

\def\mP{{\bm{P}}}
\def\mQ{{\bm{Q}}}

\def\mV{{\bm{V}}}
\def\mW{{\bm{W}}}

\def\mLam{{\bm{\Lambda}}}

\DeclareMathAlphabet{\mathsfit}{\encodingdefault}{\sfdefault}{m}{sl}
\SetMathAlphabet{\mathsfit}{bold}{\encodingdefault}{\sfdefault}{bx}{n}

\def\gK{{\mathcal{K}}}

\def\sR{{\mathbb{R}}}

\newtheorem{example}{\textbf{Example}}[section]

\title{QuadEnhancer: Leveraging Quadratic Transformations to Enhance Deep Neural Networks}

\usepackage{authblk}

\renewcommand{\thefootnote}{\fnsymbol{footnote}}

\author[1,2]{Qian Chen}
\author[2,3]{Linxin Yang}
\author[2,3,*]{Akang Wang}
\author[2]{Xiaodong Luo}
\author[3,*]{Yin Zhang}

\affil[1]{School of Science and Engineering, The Chinese University of Hong Kong, Shenzhen, China }
\affil[2]{Shenzhen International Center for Industrial and Applied Mathematics, Shenzhen Research Institute of Big Data, China}
\affil[3]{School of Data Science, The Chinese University of Hong Kong, Shenzhen, China}

\begin{document}

\maketitle
\renewcommand{\thefootnote}{*}
\footnotetext{Corresponding authors: Yin Zhang \textless yinzhang@cuhk.edu.cn \textgreater, Akang Wang \textless wangakang@sribd.cn\textgreater}

\begin{abstract}
    The combination of linear transformations and non-linear activation functions forms the foundation of most modern deep neural networks, enabling them to approximate highly complex functions. This paper explores the introduction of quadratic transformations to further increase nonlinearity in neural networks, with the aim of enhancing the performance of existing architectures. To reduce parameter complexity and computational complexity, we propose a lightweight quadratic enhancer that uses low-rankness, weight sharing, and sparsification techniques. For a fixed architecture, the proposed approach introduces quadratic interactions between features at every layer, while only adding negligible amounts of additional model parameters and forward computations. We conduct a set of proof-of-concept experiments for the proposed method across three tasks: image classification, text classification, and fine-tuning large-language models. In all tasks, the proposed approach demonstrates clear and substantial performance gains.
\end{abstract}

\section{Introduction}

In modern deep learning, the majority of successful architectures are built on the combination of linear transformations followed by non-linear activation functions. This fundamental framework allows neural networks to learn complex mappings from input to output. The linear transformation serves to project the input data into a different space, while the non-linear activation function introduces the necessary complexity, enabling the network to model intricate, high-dimensional patterns. 

The introduction of non-linearity has been a key factor in the success of neural networks, enabling them to approximate highly complex functions and capture intricate data patterns. This ability is what allows neural networks to tackle problems ranging from image classification to natural language processing. Over the years, significant advancements have been made in the architecture of neural networks, starting with MLPs for simple regression and classification tasks \citep{rumelhart1986learning,bishop1995neural}, followed by CNNs for image data \citep{lecun1989backpropagation,krizhevsky2012imagenet,he2016deep,li2021survey}, RNNs and LSTMs for sequential data \citep{hochreiter1997long,sutskever2014sequence,lipton2015critical,lindemann2021survey}, and more recently, Transformer models \citep{vaswani2017attention} that dominate a wide range of fields, including natural language processing \citep{radford2018improving,devlin2019bert,patwardhan2023transformers}, computer vision \citep{khan2022transformers,dosovitskiy2020image}, and beyond \citep{velivckovic2017graph,irwin2022chemformer}.

The success of these architectures highlights the importance of incorporating non-linearity to achieve the level of model expressiveness needed for complex real-world tasks. As a result, researchers have continually sought ways to introduce more advanced non-linear transformations within neural networks to enhance their capabilities. These explorations have generally followed three main directions: (i) employing more complex activation functions, (ii) designing non-linear network modules, and (iii) replacing linear operations with polynomial transformations.

Firstly, recent advancements in neural network activation functions have focused on introducing more complex forms of non-linearity to improve model expressiveness and tackle real-world tasks. For instance, Swish \citep{ramachandran2017swish}, which combines a sigmoid function with a linear term, offers a smoother, more flexible non-linearity, outperforming ReLU in deep networks. Further enhancements include Mish \citep{misra2019mish}, which introduces a combination of tanh and softplus to provide even finer gradients, and GELU \citep{hendrycks2016gaussian}, a probabilistic function used in Transformer models, which incorporates multiple non-linear operations such as tanh and cubic terms to introduce smooth, complex non-linearity. Despite their success, all these activation functions focus on element-wise transformations, capturing non-linearity at the level of individual neurons, yet often fail to exploit the potential for interactions between neurons that could further enhance representational capacity.

Secondly, there have been attempts to design non-linear network modules that go beyond the simple application of activation functions. For example, the introduction of GRU \citep{cho2014learning} and LSTM \citep{hochreiter1997long} units in recurrent networks adds non-linearity by incorporating gates that regulate the flow of information over time. Additionally, the attention mechanism \citep{vaswani2017attention}, introduces context-dependent weighting of neurons, allowing the network to focus on more relevant parts of the input. While these modular approaches have shown great success, they are often task-specific and depend on network architecture, making them less universally applicable.

Finally, replacing linear operations with polynomial transformations represents a more radical approach to non-linearity. This approach includes polynomial networks \citep{chrysos2021deep}, which replace standard linear layers with polynomial operations and reduce the number of parameters needed for higher-order terms through tensor decomposition techniques. Recently, \cite{mantini2021cqnn} extended second-order methods to convolutional neural networks. \cite{fan2023one,fan2023expressivity} explored the advantages of quadratic transformations in terms of both representational power and training efficiency. More recently, \cite{xu2024quadranet,xu2024quadranetv2} combined quadratic methods with neural architecture search, further improving model performance. These methods have demonstrated both theoretically and experimentally that higher-order transformations can enhance a model's representational capacity, thereby boosting its overall performance.
However, these approaches have typically been limited due to the substantial increase in parameters and computational cost associated with higher-order terms.

While methods involving more complex activation functions and specialized network modules have demonstrated significant success, polynomial transformations—despite their theoretical potential—have been underexplored in many real-world applications. The main challenge lies in the fact that higher-order terms require a considerable number of parameters, which can significantly increase model complexity. For instance, even with aggressive decomposition techniques, the work of~\cite{chrysos2021deep} still requires $O(n^2)$ parameters in high-order terms. Nevertheless, as a complementary approach to existing methods, higher-order transformations offer the potential to further enhance expressiveness without conflicting with activation functions or specialized network modules. This motivates the investigation of how polynomial transformations can be integrated with standard non-linearities to improve model performance while maintaining control over computational complexity.

In this paper, we propose a novel enhancement to traditional neural network architectures by introducing a quadratic transformation at each linear layer. This quadratic enhancement introduces higher-order interactions between neurons through quadratic terms, while maintaining computational efficiency by reusing the linear activation outputs and sparsifying parameter matrix of the quadratic term. Our method significantly reduces the number of parameters and operations required for standard quadratic transformations, making it light-weight, easily applicable to modern architectures. The key contributions of this paper are as follows:

\begin{itemize}
    \item We introduce a novel quadratic transformation technique that enhances non-linearity in neural networks by leveraging quadratic transformations to capture richer interactions between neurons.
    \item We present a sparsified version of the quadratic transformation that reduces the number of parameters, ensuring that the additional computational overhead remains minimal.
    % \item We provide a detailed cost analysis that demonstrates the computational efficiency of our approach.
    \item We evaluate the effectiveness of our approach through extensive experiments on three tasks, including image classification, text classification, and fine-tuning of large language models (LLMs), showing substantial performance improvements over baseline models.
\end{itemize}

\section{Preliminaries}

\subsection{Notation} 
Unless otherwise specified, scalars are denoted by normal font (e.g., $x$, $y$), vectors are denoted by bold lowercase letters (i.e., $\vx$, $\bm{\lambda}$), and matrices are denoted by bold uppercase letters (e.g., $\mW$, $\mP$ and $\mLam$).

\subsection{Standard linear transformation}
A typical linear transformation converts an input signal $\vx \in \sR^n$ to a feature vector $\vy \in \sR^d$ by multiplying a weight matrix $\mW \in \sR^{d \times n}$ as
\begin{equation}
    \vy := \mW \vx + \vb.
\label{eq:linear_transform}
\end{equation}
The transformed vector $\vy$ is then fed to non-linear activation functions. When stacked together in multiple layers, such linear operations, combined with nonlinear activation functions, form the basis for more complex neural network architectures. These layers can be repeated and organized in various ways to tackle a wide range of machine learning tasks, from classification to regression, progressively transforming input data into high-level abstractions.

\subsection{Primary objective}

The goal of this work is to replace the standard linear operation in a neural network layer with a quadratic function, while keeping activation functions unchanged. Specifically, we aim to replace the linear transformation (\ref{eq:linear_transform}) with a quadratic function $g:\sR^n \rightarrow \sR^d$, such that the output becomes:
\begin{equation}
    \vz=g(\vx;\mW,\mLam)
\end{equation}
where $\mW$ still represents the parameters of the linear term, and $\mLam$ represents the parameters associated with the quadratic terms. A key challenge is to ensure that the additional matrix $\mLam$, responsible for quadratic terms, only adds a small number of extra parameters relative to the existing ones in $\mW$, while the extra computational cost introduced by the quadratic transformation $g$ remains minimal compared to the standard linear transformation. We seek to enhance the expressiveness of the model by introducing higher-order interactions between features without significantly increasing the model's complexity or computational overhead.

\section{Methodologies}

Based on the objective outlined in Section 2.3, this section will provide a detailed description of the design of the quadratic function $g(\vx;\mW,\mLam)$.

\subsection{Quadratic transformation in a single layer}
Let us begin by considering a standard quadratic transformation that introduces additional nonlinearity to the linear transformation in Equation (\ref{eq:linear_transform}) by adding a quadratic term. The resulting transformation is given by:
\begin{equation}\label{eq:quad_transform}
    \vz := \begin{bmatrix}\vx^\top \mV_1 \vx\\\vdots \\ \vx^\top \mV_d \vx\end{bmatrix} + \mW \vx + \vb,
\end{equation}
where $\mV_1,\dots,\mV_d \in \sR^{n\times n}$ are the trainable weights for the quadratic terms, $\mW \in \sR^{d \times n}$  represents the weights for the linear term, and $\vb\in\sR^d$ is the bias vector. However, as is, this modification introduces a significant increase in the number of parameters, requiring $(dn^2)$ additional parameters, which would substantially increase the complexity of the model. 

\subsection{Rank-1 matrices for quadratic terms}
A common technique to reduce the number of parameters in matrices is to impose low-rankness. Specifically, we set each matrix $\mV_i,\forall i=1,\dots,d$, in (\ref{eq:quad_transform}) to be rank-1.  That is, for two vectors $\vp_i,\vq_i \in \sR^n$, 
\begin{equation}
    \mV_i:=\vp_i \vq_i^\top, ~~~\forall~ i = 1, \dots, d.
\end{equation}
 Consequently, Equation (\ref{eq:quad_transform}) becomes
 \begin{equation}\label{eq:quad_transform2}
 \vz = (\mP \vx)\odot (\mQ \vx) + \mW \vx + \vb, 
 \end{equation}
 where $\mP=\left[\vp_1,\cdots,\vp_d\right]^\top \in \sR^{d \times n}$, $\mQ=\left[ \vq_1,\dots,\vq_d\right]^\top \in \sR^{d \times n}$ and $\odot$ denotes the Hadamard (element-wise) product. This approach effectively reduces the number of extra parameters from $(dn^2)$ to $(2dn)$, significantly lowering the model's complexity while still allowing it to capture the quadratic interactions between the input features.

\subsection{Weight sharing}
To further reduce the computational and parameter complexities of the quadratic transformation, we introduce weight sharing. Specifically, we share the weight matrix $\mW$ between the linear and quadratic terms. By defining:
\begin{equation}
    \mP:=\mLam \mW, ~~~ \mQ:=\mW,
\end{equation} where $\mLam \in \sR^{d\times d}$ is a new weight matrix that differentiates the feature space of $\mP$ and $\mQ$, we can rewrite the quadratic transformation as:
\begin{equation}
    \vz = (\mLam \mW \vx) \odot (\mW \vx) + \mW \vx + \vb = (\mLam \tilde{\vy}) \odot \tilde{\vy} + \tilde{\vy} + \vb,
\end{equation}
where $\tilde{\vy}:=\mW \vx$ represents the linear transformation of the input.
This reuse of the weight matrix $\mW$ provides two key advantages. First, by sharing $\mW$ across both the linear and quadratic components, we significantly reduce the number of model parameters. Instead of three independent parameters $\mW, \mP, \mQ \in \sR^{d \times n}$, the model now only needs to learn $\mW$ and $\mLam$. Second, the linear response $\tilde{\vy}=\mW\vx$ is computed once and reused in both the linear and quadratic operations, reducing the computational overhead.

\subsection{Sparsification of $\mathbf{\Lambda}$}
\begin{wrapfigure}[11]{R}{0.5\textwidth}\vspace{-2em}
    \begin{center}
        \includegraphics[width=1\linewidth]{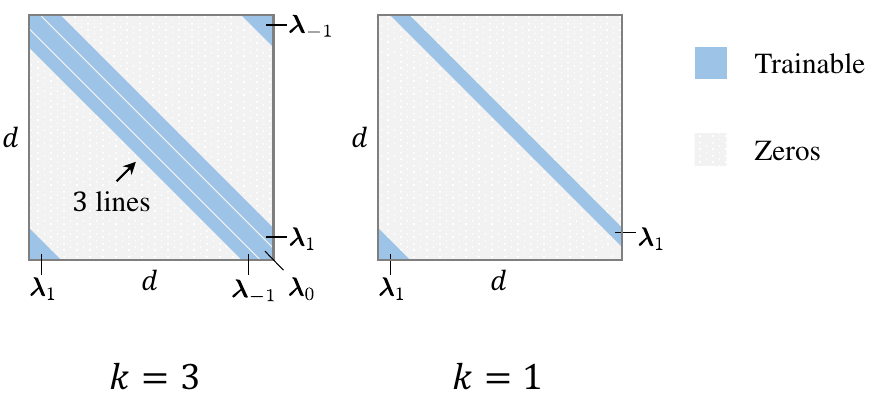}
        \vspace{-1em}
        \caption{Sparse structure of $\mLam$.}
        \label{fig:sparse_lambda}
    \end{center}
\end{wrapfigure}
While the rank-1 decomposition significantly reduces the number of parameters in the quadratic term, the weight matrix $\mLam$ still requires $O(d^2)$ parameters, which could result in a substantial increase in model complexity. To address this, we apply a sparsification strategy to $\mLam$ by converting it into a band matrix. In this structure, non-zero elements are restricted to a specific "band".
Additionally, we introduce two small triangular regions in the lower-left and upper-right corners, as illustrated in Figure \ref{fig:sparse_lambda}. This allows the band matrix with width $k$ to be divided into $k$ lines of $d$-dimensional parameters, where the missing elements of these lines are filled by values from the triangular regions. Hence, the total number of trainable parameters in $\mLam$ is reduced to $k \times d$, with $k$ being much smaller than $d$ (e.g., $k=1$). With this sparse structure, the computation of $\mLam \tilde{\vy}$ becomes:
\begin{equation}
    \mLam \tilde{\vy} = \sum_{r\in \gK} \bm{\lambda}_r \odot \text{Roll}(\tilde{\vy},r),
\end{equation}
where $\gK:=\{\cdots,-1,0,1,\cdots\}$ is the set of shifts with $|\gK|=k$, $\bm{\lambda}_r$ represents a line of parameters in $\mLam$, and the function $\text{Roll}(\cdot)$ is defined as
\begin{equation}
    \text{Roll}(\tilde{\vy},r):=[\tilde{y}_{1 + (r\bmod d)}, \dots, \tilde{y}_{d + (r\bmod d)}]^\top.
\end{equation}
The advantage of using a band matrix is that it ensures the rank of $\mLam$ remains sufficiently high, even when the number of parameters is significantly reduced (e.g., when $k=1$). Specifically, it guarantees that $\text{rank}(\mP)\leq\min\{\text{rank}(\mLam),\text{rank}(\mW)\}$ does not get too small, preserving the representational capacity of the quadratic term. This is crucial for maintaining the expressiveness of the model while reducing both parameter count and computational overhead.
In our experiments, we exclude the shift $r=0$, as it produces square terms $\tilde{\vy}^2$, whereas any non‑zero shift produces cross terms. In practice, the square terms are more prone to numerical instabilities (such as overflows or exploding gradients) than the cross terms, especially when training with the FP16 precision whose limited dynamic range amplifies numerical instability issues. The rationale behind this design choice is illustrated in Example \ref{example:pure_vs_cross}. Consequently, our experiments use $\gK = \{1\}$, as shown in the right-hand subfigure of Figure \ref{fig:sparse_lambda}. This choice effectively results in quadratic interactions between nearest-neighbor neurons, excluding the self-interaction.

\begin{example}
    Let $x_{1},x_{2}\stackrel{\text{i.i.d.}}{\sim}\mathcal{N}(0,1)$ be independent standard normal random variables. The expectations and variances of the square terms $x_1^2 $ and $x_2^2$, as well as the cross term $x_1x_2$ are as follows: $E[x_1^2] = E[x_2^2] = 1, Var[x_1^2] = Var[x_2^2] = 2$ and  $E[x_1 x_2] = 0, Var[x_1 x_2] = 1$. The cross term retains the expectations and variances of a normal distribution, while the square terms do not. Monte Carlo estimates in Table \ref{tab:prob_cross_term} further show that the square terms are far more likely to attain large absolute values compared to the cross terms.
    \begin{table}[htbp]
    \centering
    \renewcommand{\arraystretch}{1.2}
    \caption{Estimated probability of the pure quadratic and cross terms}
    \begin{tabular}{lccc}
    \hline
   probability  & $v = 4$ & $v = 8$ & $v = 16$ \\ \hline
    $p(|x_1^2| > v)$ & $4.5 \times 10^{-2}$ & $4.7 \times 10^{-3}$ & $6.33 \times 10^{-5}$ \\
    $p(|x_1 x_2| > v)$ & $7.6 \times 10^{-3}$ & $4.1\times 10^{-5}$ & $3.37 \times 10^{-10}$ \\ \hline
    \end{tabular}
    \label{tab:prob_cross_term}
    \end{table}

\label{example:pure_vs_cross}
\end{example}

\subsection{Quadratic enhancer}

\begin{figure}[h]
    \centering
    \includegraphics[width=1\linewidth]{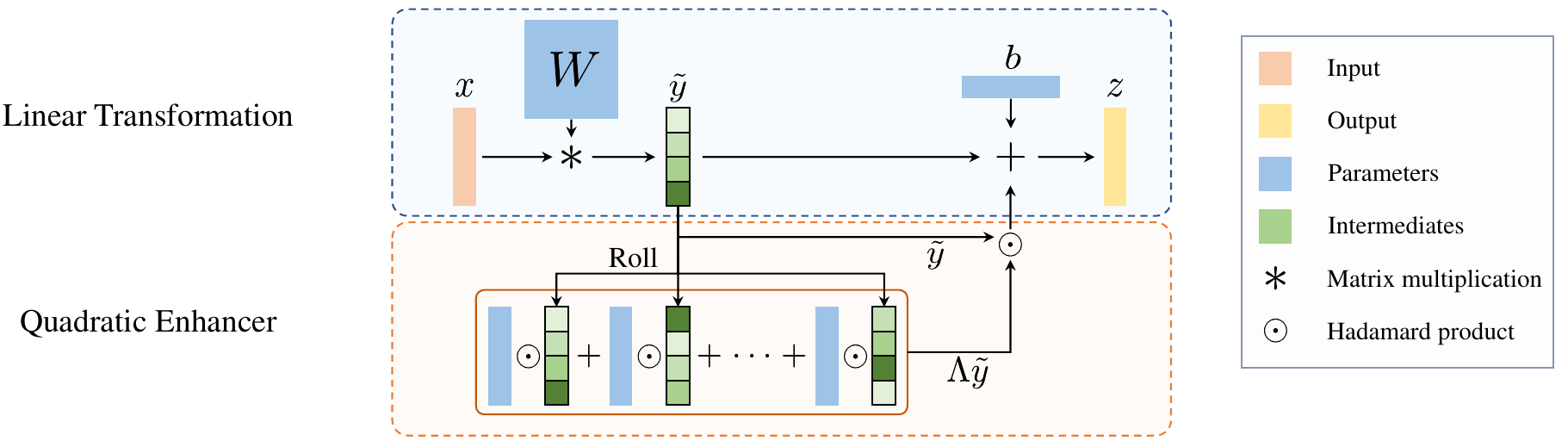}
    \caption{An overview of the quadratic enhancer.}
    \label{fig:overview}
    \vspace{-0.3cm}
\end{figure}

The complete workflow of the proposed quadratic enhancer is sketched in Figure \ref{fig:overview}, which is divided into two panels.
We recall the conventional linear transformation with bias $\vz=\mW \vx + \vb$ in the upper panel. The lower panel then details the quadratic enhancer itself. First, the linear transformation $\tilde{\vy}=\mW \vx$ is computed and fed to the enhancer. A set of rolling shifts $\text{Roll}(\cdot)$ is applied to $\tilde{\vy}$; these shifted copies are linearly combined by a learnable, band‑sparse matrix $\mLam$, producing the refined response $\mLam \tilde{\vy}$, after which a quadratically augmented feature $(\mLam \tilde{\vy}) \odot \tilde{\vy}$ is calculated. The final output is then $\vz = (\mLam \tilde{\vy}) \odot \tilde{\vy} + \tilde{\vy} + \vb$.
Figure \ref{fig:overview} illustrates how a single linear transformation can be augmented by the quadratic enhancer. Crucially, the same enhancer block can be attached to every linear layer in a neural network, endowing the entire model with richer quadratic interactions while incurring only a negligible increase in parameters and computation.

\subsection{Cost analysis}

In this section, we evaluate the overhead introduced by the quadratic enhancement in terms of both parameters and inference FLOPs, which ultimately impact the model's efficiency.

\textbf{Parameters:} The quadratic enhancer introduces a single learnable matrix $\mLam \in \sR^{d \times d}$, but with non-zero entries restricted to a bandwidth of width $k$ (where $k\ll d$, e.g., $k=1$).  This results in $k\times d$ free parameters in $\mLam$. In contrast, the standard linear layer employs a weight matrix $\mW \in \sR^{d \times n}$ with $d\times n$ parameters. The relative overhead in terms of parameters is thus: $\frac{kd}{nd}=O(k/n)$, which is negligible in practice given that $k\ll n$ (typically $n\approx d$).

\textbf{Inference FLOPs:} The quadratic enhancer reuses the linear activation $\tilde{\vy}=\mW \vx$, resulting in only three additional operations. These include: (i) a matrix multiplication $\mLam \tilde{\vy}$ which requires $2kd$ FLOPs, (ii) a Hadamard product $\odot \tilde{\vy}$ which costs $d$ FLOPs, and (iii) an addition $+\tilde{\vy}$ which adds another $d$ FLOPs. Therefore, the total number of extra FLOPs introduced by the quadratic enhancer is $2(k+1)d$ FLOPs. For comparison, the original linear transformation $\mW \vx$ requires $2nd$ FLOPs, and the bias addition adds $d$ FLOPs, leading to a total of $2nd+d$ FLOPs for the original computation. The relative overhead is thus: $\frac{2(k+1)d}{2nd+d}=O(k/n)$, which, again, becomes negligible for $k\ll n$.

\section{Experimental results}

In this section, we conduct an evaluation of the proposed quadratic enhancer across three tasks. To investigate the performance with minimal overhead, we focus on evaluating the quadratic enhancer with $\gK=\{1\}$. All experiments were conducted using four NVIDIA A100 80GB.
The experimental code is publicly available at \url{https://github.com/chitar/QuadEnhancer}.

\subsection{Image classification}

Image classification represents one of the most classical and foundational tasks in computer vision. It serves as a benchmark for evaluating the effectiveness of novel neural architectures and training methodologies.
Following common practice, our experiments involve initial pre-training on a large-scale dataset, subsequently fine-tuning the pre-trained models on various target datasets. Specifically, we first pre-train on ImageNet-1k \citep{krizhevsky2009learning}, then fine-tune and evaluate the models across several diverse downstream datasets.

\textbf{Datasets:} Our experiments begin with ImageNet-1k for the initial pre-training stage. For downstream evaluation, we use six widely recognized benchmarks: Caltech \citep{fei2004learning}, CIFAR-10, CIFAR-100 \citep{krizhevsky2009learning}, Flowers \citep{nilsback2008automated}, Food \citep{bossard2014food}, and Pets \citep{parkhi2012cats}. Caltech is a classical image recognition dataset containing various object categories. CIFAR-10 and CIFAR-100 are commonly used for small-scale visual recognition, with the former covering 10 categories and the latter 100. The Flowers dataset focuses on fine-grained classification of flower species, while the Food dataset includes a variety of food categories. The Pets dataset is centered on distinguishing different pet breeds. Further details about these datasets can be found in Table \ref{tab:img_datasets}. 

\begin{wraptable}[9]{R}{0.35\textwidth}
\centering
\vspace{-2.0em}
\caption{Image datasets}
\vspace{1em}
\scalebox{0.85}{
\begin{tabular}{lcc}
\toprule
Dataset & \# Classes & \# Samples \\
\midrule
caltech & 102 & 9.0k \\
cifar10 & 10 & 60k \\
cifar100 & 100 & 60k \\
flower & 102 & 8.1k \\
pet & 37 & 7.4k \\
food & 101 & 101k \\
\bottomrule
\end{tabular}
}
\label{tab:img_datasets}
\end{wraptable}

\textbf{Baselines:} The Vision Transformer (ViT) \citep{dosovitskiy2020image} serves as our baseline model due to its demonstrated effectiveness and wide adoption. By varying key hyperparameters such as hidden size and number of layers, we consider three model sizes: ViT-M, ViT-XT, and ViT-T. Specific hyperparameter configurations are detailed in the Table \ref{tab:vit-params}.

\begin{wraptable}[8]{R}{0.40\textwidth}
\vspace{-1.5em}
    \centering
    \caption{Model parameters of ViTs.}
    \vspace{1em}
    \scalebox{0.85}{
    \begin{tabular}{lccc}
    \toprule
     & Embed\_dim & Layers & FFN\_dim \\ 
    \midrule
    ViT-M & 192 & 6 & 768 \\
    ViT-XT & 128 & 12 & 512 \\
    ViT-T & 192 & 12 & 768 \\
    \bottomrule
    \end{tabular}
    }
    \label{tab:vit-params}
\end{wraptable}

\textbf{Training settings:} In the pre-training phase, each baseline model and its corresponding quadratic enhancer variant are trained on ImageNet-1k. The training parameters, including batch size, learning rate, number of epochs, and total training duration, are consistent with the settings outlined in \cite{mehta2022cvnets}. After pre-training, each model is fine-tuned individually on all downstream datasets, with performance evaluated on the corresponding validation sets.

\textbf{Results:} The experimental results are presented in Table \ref{tab:vit_quadratic_enhancer}, where the column labeled "ImageNet" indicates validation accuracy obtained during pre-training, and subsequent columns represent accuracy scores achieved on each downstream dataset after fine-tuning. Models incorporating the quadratic enhancer are indicated with "+QE." As shown in Table \ref{tab:vit_quadratic_enhancer}, models equipped with the quadratic enhancer consistently outperform their baseline counterparts across all datasets. Specifically, ViT-M+QE surpasses ViT-M by $1.60\%$ on ImageNet and achieves substantial gains on downstream tasks, notably improving Caltech by $2.55\%$ and CIFAR-100 by $2.34\%$. Similarly, ViT-XT+QE and ViT-T+QE outperform their respective baselines, demonstrating significant accuracy boosts across datasets. The quadratic enhancer proves especially beneficial on the challenging Pets dataset, where ViT-XT+QE achieves an impressive $6.94\%$ improvement over ViT-XT. Overall, these consistent performance enhancements underline the quadratic enhancer’s capability to enrich model effectiveness across various visual classification tasks.

\begin{table}[htbp]
\centering
\renewcommand{\arraystretch}{1.15}
\caption{Accuracy (\%) of ViTs with and without the quadratic enhancer.}
\scalebox{0.85}{
\begin{tabular}{lcccccccccc}
\toprule
\textbf{Model} & \textbf{Params} & \textbf{ImageNet} & \textbf{Caltech} & \textbf{Cifar10} & \textbf{Cifar100} & \textbf{Flowers} & \textbf{Food} & \textbf{Pets} & \textbf{Avg} \\
\midrule
ViT-M            & 2.45M & 63.70 & 87.77 & 96.35 & 80.25 & 74.01 & 83.94 & 91.03 & 82.44 \\
ViT-M+QE   & 2.47M & \textbf{65.30} & \textbf{90.32} & \textbf{97.09} & \textbf{82.59} & \textbf{75.58} & \textbf{84.63} & \textbf{91.88} & \textbf{83.91} \\
\midrule
ViT-XT             & 2.82M & 66.04 & 90.25 & 96.51 & 81.24 & 84.41 & 85.47 & 91.03 & 84.99 \\
ViT-XT+QE      & 2.83M & \textbf{67.34} & \textbf{90.77} & \textbf{96.78} & \textbf{82.64} & \textbf{86.37} & \textbf{85.85} & \textbf{97.97} & \textbf{86.82} \\
\midrule
ViT-T              & 5.37M & 73.96 & 93.07 & 97.97 & 86.13 & 86.56 & 88.42 & 93.87 & 88.57 \\
ViT-T+QE       & 5.40M & \textbf{75.15} & \textbf{94.03} & \textbf{98.03} & \textbf{86.88} & \textbf{87.25} & \textbf{88.59} & \textbf{94.95} & \textbf{89.27} \\
\bottomrule
\end{tabular}
}
\label{tab:vit_quadratic_enhancer}
\end{table}

\subsection{Text classification}

Text classification is a cornerstone task in natural language processing, underpinning applications ranging from sentiment analysis to topic categorization. It remains a classical and fundamental benchmark for evaluating advances in language modeling and fine-tuning techniques.
Analogous to image classification, modern text classification pipelines typically involve pre-training a general-purpose language model on large corpora, followed by fine-tuning on specific downstream datasets. In our experiments, we adhere to this convention: we first pre-train on a small-scale corpus, then fine-tune on several text classification benchmarks.

\textbf{Datasets:} For pre-training, we use the WikiText-2 dataset \citep{merity2016pointer}, a widely adopted corpus containing over 2 million tokens from English Wikipedia articles. While larger pre-training datasets exist, their computational demands exceed our resource constraints. For downstream text classification, we utilize six standard benchmarks: IMDB (movie review sentiment analysis) \citep{maas2011learning}, Yelp (restaurant review sentiment) \citep{yelp_dataset}, AG-News (topic classification) \citep{zhang2015character}, SST-2 (Stanford Sentiment Treebank) \citep{socher2013recursive}, and Emotion (emotion recognition) \citep{saravia-etal-2018-carer}. Detailed dataset statistics are available in Table \ref{tab:text_datasets}.

\begin{wraptable}[8]{R}{0.35\textwidth}
\vspace{-1.5em}
\centering
\caption{Text datasets.}
\vspace{1em}
\scalebox{0.85}{
\begin{tabular}{lcc}
\toprule
Dataset & \# Classes & \# Samples \\
\midrule
IMDB & 2 & 50k \\
Yelp & 5 & 700k \\
AG-News & 4 & 120k \\
SST-2 & 2 & 67k \\
Emotion & 6 & 20k \\
\bottomrule
\end{tabular}
}

\label{tab:text_datasets}
\end{wraptable}

\textbf{Baselines:} We use the GPT-2 architecture \citep{radford2019language} as our baseline, given its strong performance in language modeling and its widespread applicability across various downstream tasks. We evaluate two models: GPT-2/16 and GPT-2/32, differentiated by the size of their hidden dimension. %Both models consist of two layers.

\textbf{Training settings:} Each model is pre-trained on WikiText-2 for 20 epochs, with batch size 128, learning rate $0.0001$, and a maximum sequence length of 256 tokens, sufficient to cover most samples. Following pre-training, models are fine-tuned on each classification dataset for 10 epochs, with learning rate $0.00005$, batch size 16, and other optimizer settings unchanged.

\textbf{Results:} Table \ref{tab:gpt2_qe} reports the perplexity on WikiText-2 and validation accuracy on each classification dataset. The first column lists model variants, with '+QE' indicating use of the quadratic enhancer. The second column shows the number of trainable parameters; the third column reports WikiText-2 perplexity (lower is better). Remaining columns present classification accuracy.
As shown in Table \ref{tab:gpt2_qe}, models augmented with the quadratic enhancer achieve consistent improvements. GPT-2/16+QE reduces perplexity from 4.90 to 4.81 and increases average accuracy by $0.91\%$, with notable gains on IMDB and Emotion datasets. GPT-2/32+QE yields a perplexity drop from 4.57 to 4.44, and boosts average accuracy by $0.86\%$, driven by significant improvements on the Emotion benchmark (from $64.50\%$ to $69.05\%$). These results demonstrate that even with minimal additional parameters, the quadratic enhancer can enhance language model expressiveness and downstream performance.

\begin{table}[H]
\centering
\renewcommand{\arraystretch}{1.15}
\caption{Performance of GPT‑2 variants with and without the quadratic enhancer (QE).
WikiText is evaluated in perplexity (lower is better), whereas all other datasets are reported in accuracy (\%).}
\scalebox{0.85}{
\begin{tabular}{lcccccccc}
\hline
\textbf{Model} & \textbf{Params} & \textbf{WikiText} (ppl.) & \textbf{IMDB} & \textbf{Yelp} & \textbf{AG‑News} & \textbf{SST‑2} & \textbf{Emotion} & \textbf{Avg Acc.} \\
\toprule
GPT‑2/16      & 0.71M & 4.90 & 79.68 & 93.51 & 90.90 & \textbf{79.70} & 39.70 & 76.70 \\
GPT‑2/16+QE   & 0.82M & \textbf{4.81} & \textbf{81.16} & \textbf{93.64} & \textbf{91.67} & 78.78 & \textbf{42.80} & \textbf{77.61} \\
\midrule
GPT‑2/32      & 1.56M & 4.57 & \textbf{84.01} & 93.72 & 91.97 & 81.53 & 64.50 & 83.15 \\
GPT‑2/32+QE   & 1.61M & \textbf{4.44} & 83.52 & \textbf{93.80} & \textbf{92.01} & \textbf{81.65} & \textbf{69.05} & \textbf{84.01} \\
\bottomrule
\end{tabular}
}
\label{tab:gpt2_qe}
\end{table}

\subsection{LLMs finetuning}

Fine-tuning LLMs through parameter-efficient fine-tuning \citep{han2024parameter} is critically important due to its capability to efficiently adapt powerful pretrained models to diverse downstream tasks with minimal additional resources. We evaluate the quadratic enhancer's integration into the LoRA algorithm \citep{hu2022lora} and fine-tune three variants of the open-source LLaMA models \citep{touvron2023llama,touvron2023llama2,grattafiori2024llama3}.

\textbf{Dataset:} Our fine-tuning experiments focus on the commonsense reasoning task, a crucial benchmark to assess language models' practical reasoning capabilities. We use several benchmark datasets including BoolQ, PIQA, SIQA, HellaSwag, WinoGrande, ARC-e, ARC-c, and OBQA. Each dataset assesses different aspects of commonsense understanding and reasoning skills. Detailed descriptions of these datasets are provided in the appendix of \citep{hu2023llm}.

\textbf{Baselines and training settings:} We select three baseline models for our experiments: LLaMA-7B \citep{touvron2023llama}, LLaMA2-7B \citep{touvron2023llama2}, and LLaMA3-8B \citep{grattafiori2024llama3}, reflecting a progression of model capabilities. We employ LoRA, a widely studied parameter-efficient fine-tuning method, known for its efficiency and simplicity. For each baseline, we apply the quadratic enhancer to LoRA adapters with two ranks (r=16 and r=32). Our training configurations follow the established settings from prior works \citep{hu2023llm,liu2024dora}. %The quadratic enhancer maintains efficiency by operating with minimal parameters ($k=1$), ensuring that both parameter counts and computational overhead remain low.

\textbf{Results:} The experimental results are summarized in Table \ref{tab:llama_qe}. The first three columns specify the model names, methods (with LoRA rank indicated after the `/' and `+QE' denoting the quadratic enhancer's use), and the number of parameters trained, respectively. Subsequent columns report accuracy on the commonsense reasoning benchmark datasets. Results for LoRA without QE are taken from prior works \citep{hu2023llm, liu2024dora}. As shown in Table \ref{tab:llama_qe}, the quadratic enhancer consistently and significantly improves performance across all model variants and benchmarks, even with half-parameter versions where the rank is 16. Notably, LLaMA2-7B with LoRA/16+QE achieves an impressive average accuracy improvement of $2.64\%$ over LoRA/32. Furthermore, LLaMA3-8B models integrated with QE outperform the baseline LoRA models, particularly on complex reasoning tasks like HellaSwag and ARC datasets. This demonstrates the quadratic enhancer's strong ability to enhance LLMs' reasoning capabilities with minimal additional computational cost and parameters.

\begin{table}[H]
	\centering
	\renewcommand{\arraystretch}{1.15}
	\caption{Accuracy (\%) of LoRA finetuning on LLaMA variants with and without the quadratic enhancer (QE).}
	\scalebox{0.72}{
	\begin{tabular}{llcccccccccc}
		\toprule
		\textbf{Model} & \textbf{Method} & \textbf{Params} & \textbf{BoolQ} & \textbf{PIQA} & \textbf{SIQA} & \textbf{HellaSwag} & \textbf{WinoG.} & \textbf{ARC‑e} & \textbf{ARC‑c} & \textbf{OBQA} & \textbf{Avg} \\
		\midrule
		\multirow{3}{*}{LLaMA‑7B}  & LoRA/32       & 53.5M & 68.90 & 80.70 & 77.40 & 78.10 & 78.80 & 77.80 & 61.30 & 74.80 & 74.73 \\
		& LoRA/16+QE    & 27.6M & \textbf{69.69} & \textbf{82.64} & \textbf{79.68} & \textbf{87.11} & 80.11 & \textbf{79.41} & 63.99 & 80.20 & \textbf{77.85} \\
		& LoRA/32+QE    & 54.3M & 69.14 & 81.06 & 77.99 & 74.00 & \textbf{81.29} & 79.33 & \textbf{64.16} & \textbf{80.80} & 75.97 \\
		\midrule
		\multirow{3}{*}{LLaMA2‑7B} & LoRA/32       & 53.5M & 69.80 & 79.90 & 79.50 & 83.60 & 82.60 & 79.80 & 64.70 & 81.00 & 77.61 \\
		& LoRA/16+QE    & 27.6M & \textbf{72.26} & \textbf{82.86} & 79.78 & 86.98 & \textbf{83.66} & \textbf{85.35} & \textbf{68.51} & \textbf{82.60} & \textbf{80.25} \\
		& LoRA/32+QE    & 54.3M & 69.63 & 82.53 & \textbf{80.24} & \textbf{90.01} & 83.03 & 83.41 & \textbf{68.51} & 80.80 & 79.77 \\
		\midrule
		\multirow{3}{*}{LLaMA3‑8B} & LoRA/32       & 54.0M & 70.80 & 85.20 & 79.90 & 91.70 & 84.30 & 84.20 & 71.20 & 79.00 & 80.79 \\
		& LoRA/16+QE    & 27.7M & 74.40 & 88.46 & 80.29 & \textbf{95.45} & 86.42 & \textbf{91.37} & \textbf{80.63} & 85.00 & 85.25 \\
		& LoRA/32+QE    & 54.7M & \textbf{74.92} & \textbf{89.44} & \textbf{81.32} & 95.02 & \textbf{87.29} & 89.85 & 79.60 & \textbf{86.20} & \textbf{85.46} \\
		\bottomrule
	\end{tabular}
}
	\label{tab:llama_qe}
\end{table}

\subsection{Additional experimental results}

In addition to the main results of the three tasks considered, we perform several further experiments to evaluate the performance and scalability of the proposed quadratic enhancer. These experiments provide further insights into its comparative advantages and practical considerations in different settings. The results of these additional experiments are reported below.

\paragraph{Comparison with quadratic baselines}

We additionally compare QuadEnhancer with two existing quadratic MLP variants: (i) \textbf{QuadraNet} \citep{xu2024quadranet}, which uses the quadratic formulation $\vy = \mW_a \vx \odot \mW_b \vx + \mW_c \vx$, where $\mW_a, \mW_b, \mW_c \in \mathbb{R}^{d \times n}$ are learnable parameters, and (ii) \textbf{SwiGLU} \citep{shazeer2020glu}, which defines the quadratic interaction as $\vy = (\mW_1 \vx) \odot \text{Sigmoid}(\mW_1 \vx) \odot (\mW_2 \vx)$, with $\mW_1, \mW_2 \in \mathbb{R}^{d \times n}$. All experiments are conducted on the image classification task using the ViT-M model as the backbone, and the number of parameters is matched by adjusting the hidden dimension $d$. The results, presented in Table \ref{tab:quad-baselines}, demonstrate that QuadEnhancer consistently outperforms both quadratic baselines across all evaluated datasets.

\begin{table}[htbp]
\centering
	\renewcommand{\arraystretch}{1.15}
	\caption{Accuracy (\%) of different quadratic methods.}
	\scalebox{0.85}{
\begin{tabular}{lcccccc}
\toprule
\textbf{Method} & \textbf{Param} & \textbf{Imagenet1k} & \textbf{Cifar10} & \textbf{Cifar100} & \textbf{Food}  & \textbf{Avg}   \\
\midrule
ViT-M+QuadraNet & 2.53M                  & 61.17               & 95.81            & 79.08             & 81.58          & 79.41          \\
ViT-M+SwiGLU    & 2.58M                  & 63.25               & 96.76            & 80.58             & 83.91          & 81.13          \\
ViT-M+QuadEnhancer        & 2.47M                  & \textbf{65.30}      & \textbf{97.09}   & \textbf{82.59}    & \textbf{84.63} & \textbf{82.40}\\
\bottomrule
\end{tabular}
}
\label{tab:quad-baselines}
\end{table}

\paragraph{Scaling behavior}

Understanding the scaling behavior of a model is essential for evaluating its effectiveness as both data and model sizes increase. While large-scale experiments were not feasible due to computational constraints, we conduct experiments on data and model sizes ranging from small to moderate scales. These experiments offer insights into the scaling behavior of the proposed quadratic enhancer. All experiments are conducted on the image classification task with varying hidden dimensions $d$ and dataset sizes $s$. As shown in Table \ref{tab:scaling-behavior}, we observe that the quadratic enhancer yields increasing performance gains as both the model and dataset sizes grow.

\begin{table}[H]
\centering
	\renewcommand{\arraystretch}{1.15}
	\caption{Accuracy (\%) of different scales.}
	\scalebox{0.85}{
\begin{tabular}{lcccc}
\toprule
 & $d=24,s=50k$ & $d=48,s=100k$ & $d=96,s=200k$ & $d=192,s=400k$ \\
\midrule
ViT       & 8.99                & 19.18                & 33.14                & 49.59                 \\
ViT+QE    & 9.06                & 19.95                & 34.03                & 50.78                 \\
\midrule
Gain      & \textbf{0.07}       & \textbf{0.77}        & \textbf{0.89}        & \textbf{1.19}        \\
\bottomrule
\end{tabular}
}
\label{tab:scaling-behavior}
\end{table}

\paragraph{Training time comparison}

To evaluate the trade-offs between computational cost and performance, we report the training times for two tasks: pretraining on ImageNet-1k (Table \ref{tab:time-pretrain}) and fine-tuning on CIFAR-100 (Table \ref{tab:time-finetuning}). The results show that while the quadratic-enhanced models take slightly longer to converge in the early stages of training, they ultimately achieve superior performance. These findings suggest that while QuadEnhancer introduces some initial overhead, it offers substantial long-term performance gains.

\begin{table}[H]
\centering
	\renewcommand{\arraystretch}{1.15}
	\caption{Accuracy (\%) of pretraining on ImageNet-1k at different time.}
	\scalebox{0.85}{
\begin{tabular}{lcccccc}
\toprule
\multirow{2}{*}{Method} & \multicolumn{6}{c}{\textbf{Time(sec.)}}                                                                      \\
\cmidrule{2-7}                        & {1k}    & {10k}   & {20k}   & {30k}   & {40k}   & {50k}   \\
\midrule
ViT                     & \textbf{22.28} & \textbf{52.09} & 58.35          & \textbf{63.51} & 65.96          & 65.97          \\
ViT+QE                  & 11.30          & 51.28          & \textbf{59.07} & 63.15          & \textbf{66.59} & \textbf{67.04}\\
\bottomrule
\end{tabular}
}
\label{tab:time-pretrain}
\end{table}

\begin{table}[H]
\centering
	\renewcommand{\arraystretch}{1.15}
	\caption{Accuracy (\%) of finetuning on CIFAR-100 at different time.}
	\scalebox{0.85}{
\begin{tabular}{lccccccc}
\toprule
\multirow{2}{*}{Method} & \multicolumn{7}{c}{\textbf{Time(sec.)}}                                                                              \\
\cmidrule{2-8}                        & {500}   & {1k}    & {2k}    & {3k}    & {4k}    & {5k}    & {6k}    \\
\midrule
ViT                     & 78.79 & \textbf{81.78} & 83.00           & 84.21 & 85.29          & 85.29          & 85.29          \\
ViT+QE                  & \textbf{79.33} & 81.07          & \textbf{83.43} & \textbf{84.49} & \textbf{85.48} & \textbf{86.15} & \textbf{86.53}\\
\bottomrule
\end{tabular}
}
\label{tab:time-finetuning}
\end{table}

\paragraph{Ablation study on $\gK$}

We conduct an ablation study to investigate the effect of different choices for the set $\gK$ in image classification tasks. Models with various $\gK$ sets and hidden dimensions are trained from scratch on the Caltech dataset using ViT-M as the backbone. The accuracy results, shown in Table \ref{tab:ablation-k}, indicate that increasing the size of $\gK$ generally improves performance. The most significant improvement occurs when moving from $\gK = \emptyset$ to $\gK = {1}$, with further increases in $\gK$ yielding diminishing returns. This suggests a clear trade-off between model complexity and performance.

\begin{table}[H]
\centering
	\renewcommand{\arraystretch}{1.15}
	\caption{Accuracy (\%) when using different $\gK$ and hidden size.}
	\scalebox{0.85}{
\begin{tabular}{ccccccc}
\toprule
\multirow{2}{*}{$\gK$} & \multicolumn{6}{c}{\textbf{Hidden dimension}}          \\
 \cmidrule{2-7}                      & 24    & 48    & 96    & 144   & 192   & 288   \\
 \midrule
\textbf{$\emptyset$}   & 47.16 & 54.02 & 58.20 & 59.87 & 60.50 & 61.13 \\
$\{1\}$                & 48.94 & 54.79 & 59.35 & 60.83 & 61.39 & 61.57 \\
$\{-1,1\}$             & 49.38 & 55.09 & 59.83 & 61.28 & 61.80 & 61.65 \\
$\{-2,-1,1,2\}$        & 49.72 & 55.16 & 60.28 & 61.39 & 61.50 & 61.76\\
\bottomrule
\end{tabular}
}
\label{tab:ablation-k}
\end{table}

\section{Limitation and conclusions}
\textbf{Limitation:} A key limitation of the quadratic enhancer is that it introduces additional computational overhead, which, while minimal in terms of parameters and FLOPs, still represents an extra cost compared to traditional linear transformations. However, our approach effectively minimizes this impact, ensuring that the performance improvements far outweigh the added complexity.
%In conclusion, the quadratic enhancer effectively enhances model performance without significant increases in computational burden. The approach demonstrates substantial performance improvements across multiple tasks, offering a lightweight solution to boost neural network capabilities.

\textbf{Conclusions:} In this work, we construct a class of quadratic enhancers to enable quadratic interactions among features at a neural network layer.  Compared to a fully-connected linear transformation, the quadratic enhancers require only negligible amounts of extra parameters and FLOPs. Our proof-of-concept experiments, conducted across multiple tasks and multiple datasets, have confirmed the potentials of the proposed approach in delivering substantial performance improvements without notably increasing model sizes.  Evidences even suggest that, in some cases, our approach may be able to significantly reduce model sizes while maintaining or even enhancing performance levels.  At this point, however, our study is still quite preliminary. More research is definitely needed in this direction to better understand the promises and limitations of the proposed approach.

\section*{Acknowledgments}

This work was supported by the National Key R\&D Program of China under grant 2023YFA1009300, and by the Shenzhen Science and Technology Program (Grant No.~GXWD20201231105722002-20200901175001001).
Qian Chen, Linxin Yang, and Akang Wang also acknowledge support from the National Natural
Science Foundation of China (Grant No.~12301416) and the Guangdong Basic and Applied Basic Research Foundation (Grant No.~2024A1515010306). Xiaodong Luo also acknowledges support from the Hetao Shenzhen-Hong Kong Science and Technology Innovation Cooperation Zone Project (No.~HZQSWS-KCCYB-2024016).

% \section*{References}
\bibliography{reference}
\bibliographystyle{plain}
%%%%%%%%%%%%%%%%%%%%%%%%%%%%%%%%%%%%%%%%%%%%%%%%%%%%%%%%%%%%
\newpage

\appendix

%%%%%%%%%%%%%%%%%%%%%%%%%%%%%%%%%%%%%%%%%%%%%%%%%%%%%%%%%%%%

\newpage
\section*{NeurIPS Paper Checklist}

\begin{enumerate}

\item {\bf Claims}
    \item[] Question: Do the main claims made in the abstract and introduction accurately reflect the paper's contributions and scope?
    \item[] Answer: \answerYes{} % Replace by \answerYes{}, \answerNo{}, or \answerNA{}.
    \item[] Justification: Please refer to Abstract and Section 1.
    \item[] Guidelines:
    \begin{itemize}
        \item The answer NA means that the abstract and introduction do not include the claims made in the paper.
        \item The abstract and/or introduction should clearly state the claims made, including the contributions made in the paper and important assumptions and limitations. A No or NA answer to this question will not be perceived well by the reviewers. 
        \item The claims made should match theoretical and experimental results, and reflect how much the results can be expected to generalize to other settings. 
        \item It is fine to include aspirational goals as motivation as long as it is clear that these goals are not attained by the paper. 
    \end{itemize}

\item {\bf Limitations}
    \item[] Question: Does the paper discuss the limitations of the work performed by the authors?
    \item[] Answer: \answerYes{} % Replace by \answerYes{}, \answerNo{}, or \answerNA{}.
    \item[] Justification:  Please refer to Section 5.
    \item[] Guidelines:
    \begin{itemize}
        \item The answer NA means that the paper has no limitation while the answer No means that the paper has limitations, but those are not discussed in the paper. 
        \item The authors are encouraged to create a separate "Limitations" section in their paper.
        \item The paper should point out any strong assumptions and how robust the results are to violations of these assumptions (e.g., independence assumptions, noiseless settings, model well-specification, asymptotic approximations only holding locally). The authors should reflect on how these assumptions might be violated in practice and what the implications would be.
        \item The authors should reflect on the scope of the claims made, e.g., if the approach was only tested on a few datasets or with a few runs. In general, empirical results often depend on implicit assumptions, which should be articulated.
        \item The authors should reflect on the factors that influence the performance of the approach. For example, a facial recognition algorithm may perform poorly when image resolution is low or images are taken in low lighting. Or a speech-to-text system might not be used reliably to provide closed captions for online lectures because it fails to handle technical jargon.
        \item The authors should discuss the computational efficiency of the proposed algorithms and how they scale with dataset size.
        \item If applicable, the authors should discuss possible limitations of their approach to address problems of privacy and fairness.
        \item While the authors might fear that complete honesty about limitations might be used by reviewers as grounds for rejection, a worse outcome might be that reviewers discover limitations that aren't acknowledged in the paper. The authors should use their best judgment and recognize that individual actions in favor of transparency play an important role in developing norms that preserve the integrity of the community. Reviewers will be specifically instructed to not penalize honesty concerning limitations.
    \end{itemize}

\item {\bf Theory assumptions and proofs}
    \item[] Question: For each theoretical result, does the paper provide the full set of assumptions and a complete (and correct) proof?
    \item[] Answer: \answerNA{} % Replace by \answerYes{}, \answerNo{}, or \answerNA{}.
    \item[] Justification: This paper does not include theoretical results.
    \item[] Guidelines:
    \begin{itemize}
        \item The answer NA means that the paper does not include theoretical results. 
        \item All the theorems, formulas, and proofs in the paper should be numbered and cross-referenced.
        \item All assumptions should be clearly stated or referenced in the statement of any theorems.
        \item The proofs can either appear in the main paper or the supplemental material, but if they appear in the supplemental material, the authors are encouraged to provide a short proof sketch to provide intuition. 
        \item Inversely, any informal proof provided in the core of the paper should be complemented by formal proofs provided in appendix or supplemental material.
        \item Theorems and Lemmas that the proof relies upon should be properly referenced. 
    \end{itemize}

    \item {\bf Experimental result reproducibility}
    \item[] Question: Does the paper fully disclose all the information needed to reproduce the main experimental results of the paper to the extent that it affects the main claims and/or conclusions of the paper (regardless of whether the code and data are provided or not)?
    \item[] Answer: \answerYes{} % Replace by \answerYes{}, \answerNo{}, or \answerNA{}.
    \item[] Justification: Please refer to Section 4.
    \item[] Guidelines:
    \begin{itemize}
        \item The answer NA means that the paper does not include experiments.
        \item If the paper includes experiments, a No answer to this question will not be perceived well by the reviewers: Making the paper reproducible is important, regardless of whether the code and data are provided or not.
        \item If the contribution is a dataset and/or model, the authors should describe the steps taken to make their results reproducible or verifiable. 
        \item Depending on the contribution, reproducibility can be accomplished in various ways. For example, if the contribution is a novel architecture, describing the architecture fully might suffice, or if the contribution is a specific model and empirical evaluation, it may be necessary to either make it possible for others to replicate the model with the same dataset, or provide access to the model. In general. releasing code and data is often one good way to accomplish this, but reproducibility can also be provided via detailed instructions for how to replicate the results, access to a hosted model (e.g., in the case of a large language model), releasing of a model checkpoint, or other means that are appropriate to the research performed.
        \item While NeurIPS does not require releasing code, the conference does require all submissions to provide some reasonable avenue for reproducibility, which may depend on the nature of the contribution. For example
        \begin{enumerate}
            \item If the contribution is primarily a new algorithm, the paper should make it clear how to reproduce that algorithm.
            \item If the contribution is primarily a new model architecture, the paper should describe the architecture clearly and fully.
            \item If the contribution is a new model (e.g., a large language model), then there should either be a way to access this model for reproducing the results or a way to reproduce the model (e.g., with an open-source dataset or instructions for how to construct the dataset).
            \item We recognize that reproducibility may be tricky in some cases, in which case authors are welcome to describe the particular way they provide for reproducibility. In the case of closed-source models, it may be that access to the model is limited in some way (e.g., to registered users), but it should be possible for other researchers to have some path to reproducing or verifying the results.
        \end{enumerate}
    \end{itemize}

\item {\bf Open access to data and code}
    \item[] Question: Does the paper provide open access to the data and code, with sufficient instructions to faithfully reproduce the main experimental results, as described in supplemental material?
    \item[] Answer: \answerYes{} % Replace by \answerYes{}, \answerNo{}, or \answerNA{}.
    \item[] Justification: We provide a code link in Section 4.
    \item[] Guidelines:
    \begin{itemize}
        \item The answer NA means that paper does not include experiments requiring code.
        \item Please see the NeurIPS code and data submission guidelines (\url{https://nips.cc/public/guides/CodeSubmissionPolicy}) for more details.
        \item While we encourage the release of code and data, we understand that this might not be possible, so “No” is an acceptable answer. Papers cannot be rejected simply for not including code, unless this is central to the contribution (e.g., for a new open-source benchmark).
        \item The instructions should contain the exact command and environment needed to run to reproduce the results. See the NeurIPS code and data submission guidelines (\url{https://nips.cc/public/guides/CodeSubmissionPolicy}) for more details.
        \item The authors should provide instructions on data access and preparation, including how to access the raw data, preprocessed data, intermediate data, and generated data, etc.
        \item The authors should provide scripts to reproduce all experimental results for the new proposed method and baselines. If only a subset of experiments are reproducible, they should state which ones are omitted from the script and why.
        \item At submission time, to preserve anonymity, the authors should release anonymized versions (if applicable).
        \item Providing as much information as possible in supplemental material (appended to the paper) is recommended, but including URLs to data and code is permitted.
    \end{itemize}

\item {\bf Experimental setting/details}
    \item[] Question: Does the paper specify all the training and test details (e.g., data splits, hyperparameters, how they were chosen, type of optimizer, etc.) necessary to understand the results?
    \item[] Answer: \answerYes{} % Replace by \answerYes{}, \answerNo{}, or \answerNA{}.
    \item[] Justification: Please refer to Section 4.
    \item[] Guidelines:
    \begin{itemize}
        \item The answer NA means that the paper does not include experiments.
        \item The experimental setting should be presented in the core of the paper to a level of detail that is necessary to appreciate the results and make sense of them.
        \item The full details can be provided either with the code, in appendix, or as supplemental material.
    \end{itemize}

\item {\bf Experiment statistical significance}
    \item[] Question: Does the paper report error bars suitably and correctly defined or other appropriate information about the statistical significance of the experiments?
    \item[] Answer: \answerNo{} % Replace by \answerYes{}, \answerNo{}, or \answerNA{}.
    \item[] Justification: The experiments are computationally intensive, and we have not yet had enough time to conduct repeated experiments to obtain error bars.
    \item[] Guidelines:
    \begin{itemize}
        \item The answer NA means that the paper does not include experiments.
        \item The authors should answer "Yes" if the results are accompanied by error bars, confidence intervals, or statistical significance tests, at least for the experiments that support the main claims of the paper.
        \item The factors of variability that the error bars are capturing should be clearly stated (for example, train/test split, initialization, random drawing of some parameter, or overall run with given experimental conditions).
        \item The method for calculating the error bars should be explained (closed form formula, call to a library function, bootstrap, etc.)
        \item The assumptions made should be given (e.g., Normally distributed errors).
        \item It should be clear whether the error bar is the standard deviation or the standard error of the mean.
        \item It is OK to report 1-sigma error bars, but one should state it. The authors should preferably report a 2-sigma error bar than state that they have a 96\% CI, if the hypothesis of Normality of errors is not verified.
        \item For asymmetric distributions, the authors should be careful not to show in tables or figures symmetric error bars that would yield results that are out of range (e.g. negative error rates).
        \item If error bars are reported in tables or plots, The authors should explain in the text how they were calculated and reference the corresponding figures or tables in the text.
    \end{itemize}

\item {\bf Experiments compute resources}
    \item[] Question: For each experiment, does the paper provide sufficient information on the computer resources (type of compute workers, memory, time of execution) needed to reproduce the experiments?
    \item[] Answer: \answerYes{} % Replace by \answerYes{}, \answerNo{}, or \answerNA{}.
    \item[] Justification: Please refer to Section 4.
    \item[] Guidelines:
    \begin{itemize}
        \item The answer NA means that the paper does not include experiments.
        \item The paper should indicate the type of compute workers CPU or GPU, internal cluster, or cloud provider, including relevant memory and storage.
        \item The paper should provide the amount of compute required for each of the individual experimental runs as well as estimate the total compute. 
        \item The paper should disclose whether the full research project required more compute than the experiments reported in the paper (e.g., preliminary or failed experiments that didn't make it into the paper). 
    \end{itemize}
    
\item {\bf Code of ethics}
    \item[] Question: Does the research conducted in the paper conform, in every respect, with the NeurIPS Code of Ethics \url{https://neurips.cc/public/EthicsGuidelines}?
    \item[] Answer: \answerYes{} % Replace by \answerYes{}, \answerNo{}, or \answerNA{}.
    \item[] Justification: The research presented in this paper adheres to the NeurIPS Code of Ethics.
    \item[] Guidelines:
    \begin{itemize}
        \item The answer NA means that the authors have not reviewed the NeurIPS Code of Ethics.
        \item If the authors answer No, they should explain the special circumstances that require a deviation from the Code of Ethics.
        \item The authors should make sure to preserve anonymity (e.g., if there is a special consideration due to laws or regulations in their jurisdiction).
    \end{itemize}

\item {\bf Broader impacts}
    \item[] Question: Does the paper discuss both potential positive societal impacts and negative societal impacts of the work performed?
    \item[] Answer: \answerNA{} % Replace by \answerYes{}, \answerNo{}, or \answerNA{}.
    \item[] Justification: This paper presents work whose goal is to advance the field of Machine
Learning. There are many potential societal consequences of our work, none of which we
feel must be specifically highlighted here.

    \item[] Guidelines:
    \begin{itemize}
        \item The answer NA means that there is no societal impact of the work performed.
        \item If the authors answer NA or No, they should explain why their work has no societal impact or why the paper does not address societal impact.
        \item Examples of negative societal impacts include potential malicious or unintended uses (e.g., disinformation, generating fake profiles, surveillance), fairness considerations (e.g., deployment of technologies that could make decisions that unfairly impact specific groups), privacy considerations, and security considerations.
        \item The conference expects that many papers will be foundational research and not tied to particular applications, let alone deployments. However, if there is a direct path to any negative applications, the authors should point it out. For example, it is legitimate to point out that an improvement in the quality of generative models could be used to generate deepfakes for disinformation. On the other hand, it is not needed to point out that a generic algorithm for optimizing neural networks could enable people to train models that generate Deepfakes faster.
        \item The authors should consider possible harms that could arise when the technology is being used as intended and functioning correctly, harms that could arise when the technology is being used as intended but gives incorrect results, and harms following from (intentional or unintentional) misuse of the technology.
        \item If there are negative societal impacts, the authors could also discuss possible mitigation strategies (e.g., gated release of models, providing defenses in addition to attacks, mechanisms for monitoring misuse, mechanisms to monitor how a system learns from feedback over time, improving the efficiency and accessibility of ML).
    \end{itemize}
    
\item {\bf Safeguards}
    \item[] Question: Does the paper describe safeguards that have been put in place for responsible release of data or models that have a high risk for misuse (e.g., pretrained language models, image generators, or scraped datasets)?
    \item[] Answer: \answerNA{} % Replace by \answerYes{}, \answerNo{}, or \answerNA{}.
    \item[] Justification: Our paper does not present any such risks.
    \item[] Guidelines:
    \begin{itemize}
        \item The answer NA means that the paper poses no such risks.
        \item Released models that have a high risk for misuse or dual-use should be released with necessary safeguards to allow for controlled use of the model, for example by requiring that users adhere to usage guidelines or restrictions to access the model or implementing safety filters. 
        \item Datasets that have been scraped from the Internet could pose safety risks. The authors should describe how they avoided releasing unsafe images.
        \item We recognize that providing effective safeguards is challenging, and many papers do not require this, but we encourage authors to take this into account and make a best faith effort.
    \end{itemize}

\item {\bf Licenses for existing assets}
    \item[] Question: Are the creators or original owners of assets (e.g., code, data, models), used in the paper, properly credited and are the license and terms of use explicitly mentioned and properly respected?
    \item[] Answer: \answerYes{} % Replace by \answerYes{}, \answerNo{}, or \answerNA{}.
    \item[] Justification: Please refer to Section 4.
    \item[] Guidelines:
    \begin{itemize}
        \item The answer NA means that the paper does not use existing assets.
        \item The authors should cite the original paper that produced the code package or dataset.
        \item The authors should state which version of the asset is used and, if possible, include a URL.
        \item The name of the license (e.g., CC-BY 4.0) should be included for each asset.
        \item For scraped data from a particular source (e.g., website), the copyright and terms of service of that source should be provided.
        \item If assets are released, the license, copyright information, and terms of use in the package should be provided. For popular datasets, \url{paperswithcode.com/datasets} has curated licenses for some datasets. Their licensing guide can help determine the license of a dataset.
        \item For existing datasets that are re-packaged, both the original license and the license of the derived asset (if it has changed) should be provided.
        \item If this information is not available online, the authors are encouraged to reach out to the asset's creators.
    \end{itemize}

\item {\bf New assets}
    \item[] Question: Are new assets introduced in the paper well documented and is the documentation provided alongside the assets?
    \item[] Answer: \answerYes{} % Replace by \answerYes{}, \answerNo{}, or \answerNA{}.
    \item[] Justification:  See the code link in Section 4.
    \item[] Guidelines:
    \begin{itemize}
        \item The answer NA means that the paper does not release new assets.
        \item Researchers should communicate the details of the dataset/code/model as part of their submissions via structured templates. This includes details about training, license, limitations, etc. 
        \item The paper should discuss whether and how consent was obtained from people whose asset is used.
        \item At submission time, remember to anonymize your assets (if applicable). You can either create an anonymized URL or include an anonymized zip file.
    \end{itemize}

\item {\bf Crowdsourcing and research with human subjects}
    \item[] Question: For crowdsourcing experiments and research with human subjects, does the paper include the full text of instructions given to participants and screenshots, if applicable, as well as details about compensation (if any)? 
    \item[] Answer: \answerNA{} % Replace by \answerYes{}, \answerNo{}, or \answerNA{}.
    \item[] Justification:  Our paper does not involve any crowdsourcing or research with human subjects.
    \item[] Guidelines:
    \begin{itemize}
        \item The answer NA means that the paper does not involve crowdsourcing nor research with human subjects.
        \item Including this information in the supplemental material is fine, but if the main contribution of the paper involves human subjects, then as much detail as possible should be included in the main paper. 
        \item According to the NeurIPS Code of Ethics, workers involved in data collection, curation, or other labor should be paid at least the minimum wage in the country of the data collector. 
    \end{itemize}

\item {\bf Institutional review board (IRB) approvals or equivalent for research with human subjects}
    \item[] Question: Does the paper describe potential risks incurred by study participants, whether such risks were disclosed to the subjects, and whether Institutional Review Board (IRB) approvals (or an equivalent approval/review based on the requirements of your country or institution) were obtained?
    \item[] Answer: \answerNA{} % Replace by \answerYes{}, \answerNo{}, or \answerNA{}.
    \item[] Justification:  Our paper does not involve any crowdsourcing or research with human subjects.
    \item[] Guidelines:
    \begin{itemize}
        \item The answer NA means that the paper does not involve crowdsourcing nor research with human subjects.
        \item Depending on the country in which research is conducted, IRB approval (or equivalent) may be required for any human subjects research. If you obtained IRB approval, you should clearly state this in the paper. 
        \item We recognize that the procedures for this may vary significantly between institutions and locations, and we expect authors to adhere to the NeurIPS Code of Ethics and the guidelines for their institution. 
        \item For initial submissions, do not include any information that would break anonymity (if applicable), such as the institution conducting the review.
    \end{itemize}

\item {\bf Declaration of LLM usage}
    \item[] Question: Does the paper describe the usage of LLMs if it is an important, original, or non-standard component of the core methods in this research? Note that if the LLM is used only for writing, editing, or formatting purposes and does not impact the core methodology, scientific rigorousness, or originality of the research, declaration is not required.
    %this research? 
    \item[] Answer: \answerNA{} % Replace by \answerYes{}, \answerNo{}, or \answerNA{}.
    \item[] Justification: No such usage.
    \item[] Guidelines:
    \begin{itemize}
        \item The answer NA means that the core method development in this research does not involve LLMs as any important, original, or non-standard components.
        \item Please refer to our LLM policy (\url{https://neurips.cc/Conferences/2025/LLM}) for what should or should not be described.
    \end{itemize}

\end{enumerate}

\end{document}